%% file: main.tex
\documentclass{article}

\usepackage[table]{xcolor}
\usepackage[preprint]{corl_2026}
\hypersetup{hypertexnames=false}

\usepackage{amsmath,amssymb,amsfonts}
\usepackage{booktabs}
\usepackage{tabularx}
\usepackage{adjustbox}
\usepackage{graphicx}
\usepackage{enumitem}
\usepackage{float}
\usepackage[section]{placeins}
\floatstyle{ruled}
\newfloat{algorithm}{t}{loa}
\floatname{algorithm}{Algorithm}
\graphicspath{{figures/}{./}}

\title{RCSP: Risk-Sensitive Conjectural Scenario Planning for Safe Dynamic Robot Navigation}
\author{
Zhengye Han \quad Quanyan Zhu\\
Department of Electrical and Computer Engineering\\
New York University\\
\texttt{\{zh3286,qz494\}@nyu.edu}
}

\newcommand{\method}{RCSP}

\newcommand{\cvar}{\mathrm{CVaR}}

\newcommand{\safe}{\mathrm{safe}}
\newcommand{\nom}{\mathrm{nom}}

\begin{document}
\maketitle
\vspace{-2em}
\begin{figure}[h]
\centering
\includegraphics[width=1\textwidth]{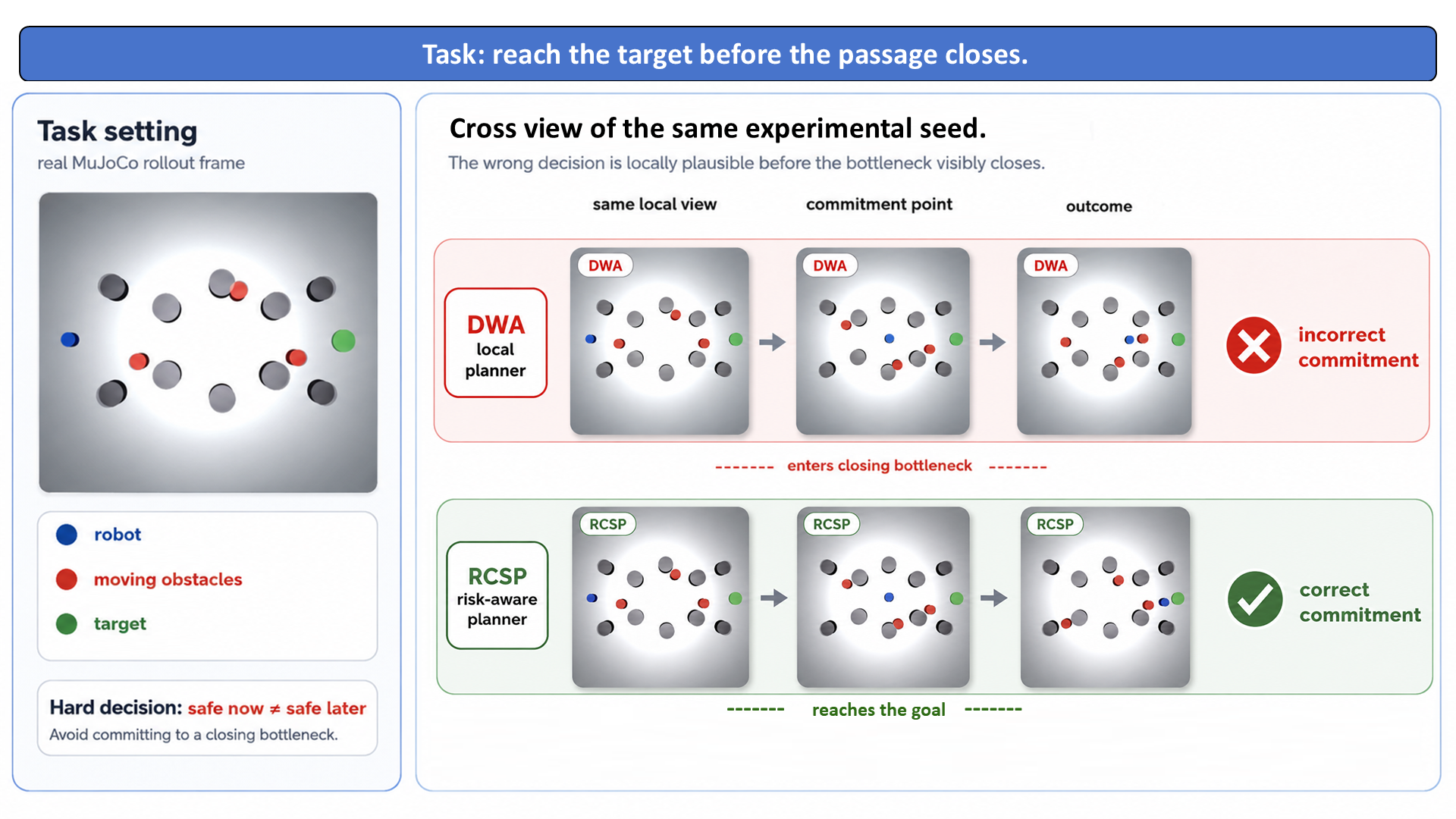}
\vspace{-1em}
\caption{Actual MuJoCo rollout frames illustrating predictive near-miss commitment. Under the same dynamic bottleneck, a DWA-style local planner commits to a short-horizon trap, while \method{} reaches the goal by scoring tail risk over plausible moving-obstacle futures.}
\label{fig:teaser}
\end{figure}

\begin{abstract}
Mobile robots can fail before they collide: a velocity that is safe now may commit the
robot to a passage that moving obstacles will soon close. We study this predictive
near-miss commitment problem and propose Risk-Sensitive Conjectural Scenario Planning
(RCSP), a planning layer that evaluates candidate commands against plausible
short-horizon obstacle futures. RCSP maintains a lightweight belief over local motion
conjectures, samples future interactions, penalizes high-risk tails, and executes
through a local safety check. In controlled MuJoCo bottleneck tasks, the RCSP planner
reaches the goal without collisions and yields higher secondary safety and path-quality
point estimates than a non-adaptive predictor, with additional latency. In ROS2/Gazebo,
adding the local safety layer to a standard Nav2 stack reduces dynamic near-miss
failures. On official DynaBARN/Jackal transfer, tuned DWA and TEB remain stronger on
strict benchmark success, revealing the boundary of the approach. These simulation
results position RCSP as a predictive-risk module that complements existing navigation
stacks in dynamic bottleneck regimes.
\end{abstract}


\section{Introduction}

Mobile robots navigating among moving obstacles must make decisions before the future is fully
visible. A velocity command can be collision-free under the current observation, improve
short-horizon progress, and still commit the robot to a doorway, corridor, or bottleneck that a
moving obstacle will occupy moments later. Once the robot has entered such a passage, the remaining
choices may be limited to braking, oscillating, or colliding. We call this failure mode
\emph{dynamic near-miss commitment}: the dangerous decision is made before the near miss appears
as an immediate local constraint violation.

Mature navigation systems, including DWA, TEB, ROS2/Nav2 controllers, velocity-obstacle methods,
MPC variants, and control-barrier-style safety filters, provide strong tools for tracking, local
obstacle avoidance, and instantaneous safety checking
\citep{fox1997dynamic,rosmann2012teb,ames2019control}. However, dynamic near-miss commitment
exposes a different weakness: local clearance and short-horizon progress can be misleading when
the risk lies in the tail of plausible future obstacle motions. This paper therefore asks whether a
robot can improve safety by evaluating candidate commands under plausible moving-obstacle futures
and penalizing high-risk tails. We study this as a controlled simulation mechanism question, using
ROS2/Gazebo near-miss tasks, MuJoCo dynamic-bottleneck environments, and official
DynaBARN/Jackal transfer, where maintained DWA and TEB remain strong public-benchmark
baselines \citep{nair2022dynabarn}.

We propose \emph{Risk-Sensitive Conjectural Scenario Planning} (\method{}), a planning-layer
mechanism that combines conjectural posterior update, CVaR scenario planning, and fixed safety
execution. At each control step, \method{} maintains a finite posterior over local obstacle-motion
conjectures, samples short-horizon futures from this posterior, scores candidate velocity commands
by a CVaR tail-risk objective \citep{rockafellar2000cvar}, and passes the selected nominal command
through a fixed control-barrier-style execution layer \citep{ames2019control}. The conjectural
update is deliberately lightweight: it is not intended to recover a true social-motion model, but to
upweight members of a small local model family that best explain the recent interaction history.

The resulting claim is targeted. In dynamic near-miss and bottleneck regimes, RCSP is designed to improve safety by
changing the robot's commitment before failure becomes locally unavoidable. On official DynaBARN/Jackal transfer, the picture is more mixed: full posterior RCSP gives better
point estimates than its fixed-predictor and lower-level ablations, but mature DWA/TEB controllers
remain stronger on official strict success. Thus, \method{} is best viewed as a predictive-risk layer to be integrated with robust
navigation stacks, not as a universal replacement for them.

Our contributions are:
\begin{enumerate}[leftmargin=1.2em,itemsep=0.2em]
    \item We identify dynamic near-miss commitment as a predictive failure mode in navigation
    around moving obstacles.
    \item We propose \method{}, a planning-layer mechanism combining online conjectural model
    updating, posterior scenario rollouts, CVaR tail-risk scoring, and fixed differential-drive
    safety execution.
    \item We evaluate \method{} across ROS2/Gazebo, MuJoCo, and official DynaBARN/Jackal transfer,
    showing both where tail-risk planning helps and where mature local planners remain stronger.
\end{enumerate}

\section{Related Work}

\paragraph{Dynamic robot navigation and local planning.}
Navigation among moving agents has a long history in local planning, social navigation, and
learning-based obstacle avoidance \citep{mavrogiannis2023core,xie2023drlvo}. DWA, TEB, and modern
ROS2/Nav2 controllers such as DWB, Regulated Pure Pursuit, Graceful Controller, and MPPI remain
strong practical baselines for local tracking, obstacle avoidance, and velocity selection
\citep{fox1997dynamic,rosmann2012teb,macenski2022robot,williams2017mppi}. BARN and DynaBARN further
stress such navigation stacks in constrained and dynamic Jackal environments
\citep{xiao2022barn,nair2022dynabarn}. RCSP is not intended to replace these systems wholesale.
Instead, it targets a narrower failure mode: local progress and instantaneous clearance can look
favorable even when the selected velocity commits the robot to a future dynamic near miss.

\paragraph{Risk-sensitive and predictive safety planning.}
Control barrier functions and MPC-CBF methods enforce local safety constraints when the relevant
state, model, and barrier conditions are well aligned \citep{ames2019control}. Predictive safety
filters extend this view by evaluating imagined, adversarial, or counterfactual futures before
execution \citep{nguyen2024gameplay}. RCSP follows this predictive-safety perspective, but uses
lightweight local rollouts and Conditional Value-at-Risk (CVaR) to penalize the high-risk tail of
plausible moving-obstacle futures \citep{rockafellar2000cvar}. Its finite-sample interpretation is
related to scenario optimization \citep{campi2008scenario}; however, RCSP uses scenarios online to
choose the next velocity command rather than offline to certify a controller.

\paragraph{Learning-based safe navigation.}
Safe reinforcement-learning methods such as CPO and PPO-Lagrangian learn constrained policies from
interaction data \citep{achiam2017cpo,ray2019benchmarking}. These methods are important comparisons
for safe decision-making, but they can require substantial task-specific training and may still incur
safety cost during training or transfer. RCSP instead remains a planning/control method: it does not
learn a task policy, adapts only a small local predictive model online, and trades additional runtime
latency for explicit scenario-based risk evaluation.

\section{Problem Formulation}

We consider a differential-drive robot navigating to a goal in a dynamic environment. At time $t$,
the robot state is $x^r_t$, the local observation is $o_t$, and the executed velocity command is
\(
    u_t=(v_t,\omega_t)\in\mathcal{U},
\)
where $v_t$ and $\omega_t$ denote linear and angular velocity. The observation $o_t$ may contain a
local range scan such as LiDAR, an occupancy crop, visible moving-obstacle tracks, or
simulator-provided local range observations. The environment contains static obstacles and moving
agents whose future states are only partially observed. The history available before acting is
\(
    \mathcal{H}_t=\{(o_\tau,u_\tau)\}_{\tau<t}\cup\{o_t\}.
\)

The failure mode studied in this paper is not an immediate collision violation, but a
\emph{predictive commitment} error. A command may be feasible and collision-free under $o_t$, yet
place the robot in a local passage whose future safety depends on uncertain moving-obstacle motion.
To reason about this uncertainty, RCSP maintains a finite conjectural model family $\Theta$ for
nearby moving-obstacle dynamics. For each $\theta\in\Theta$, $b_t^\theta$ denotes the obstacle-state
belief under conjecture $\theta$, and $q_t\in\Delta(\Theta)$ is the posterior weight over
conjectures. The robot information state is
\(
    I_t=\bigl(M_t,\{b_t^\theta\}_{\theta\in\Theta},q_t,g\bigr),
\)
where $M_t$ is the local map or occupancy memory and $g$ is the goal. Given $I_t$, the planner
chooses a velocity command before the relevant future obstacle motion is fully observed. The
objective is to reach $g$ while avoiding collisions, maintaining clearance, and not committing to
passages that are likely to close.

\section{Method}
\label{sec:method}

RCSP separates predictive risk reasoning from low-level execution. Before acting, the planner scores
available velocity commands by their tail outcomes over plausible future moving-obstacle
interactions, while a fixed safety filter enforces local feasibility and immediate safety.

\begin{figure*}[t]
\centering
\makebox[\textwidth][c]{%
\includegraphics[width=1\textwidth]{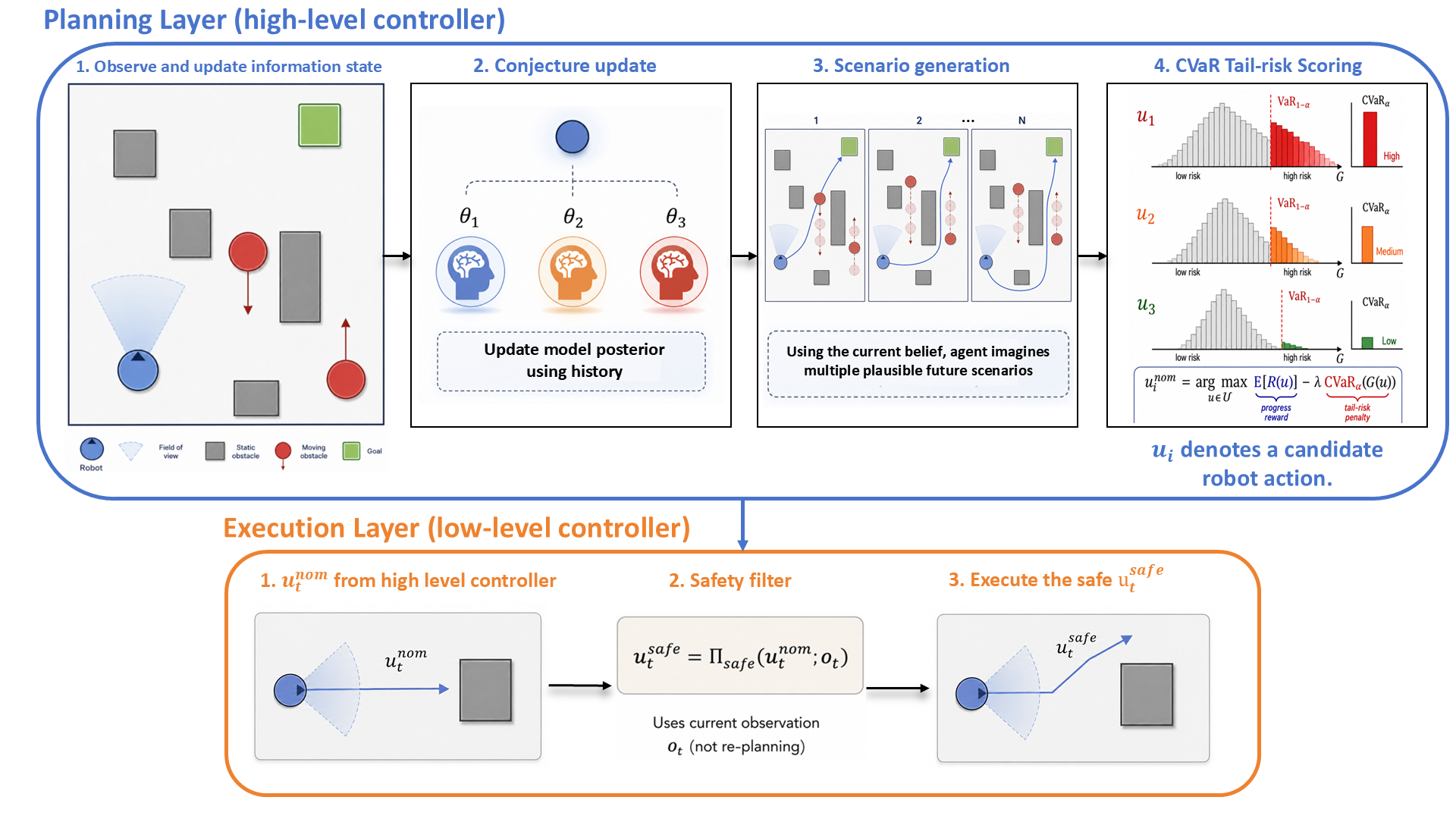}
}
\vspace{-1.5em}
\caption{Method overview. The high-level planning layer updates the information state, maintains a
posterior over local moving-obstacle conjectures, samples multiple future scenarios, and scores
candidate actions by CVaR tail risk. The low-level execution layer applies a fixed safety filter to
the planner-selected velocity command.}
\label{fig:method-flowchart}
\vspace{-1em}
\end{figure*}

\subsection{Conjectural Model Update}

The conjecture variable represents how nearby dynamic agents may move over the short horizon.
RCSP maintains posterior weights $q_t(\theta)$ over a finite model family and updates them by
predictive likelihood. This update should be interpreted as lightweight misspecified-model
selection: a highly weighted conjecture need not be the true environment model, but it should be
useful for predicting the recent local interaction history.

In our experiments, $\Theta$ contains simple local motion models for nearby moving obstacles,
including static, slow/constant/fast velocity, yielding, and aggressive variants. This family is
not intended as a complete social-motion model. Its role is to span locally plausible futures whose
weights can change online as observations arrive. The fixed-predictor ablation disables this update
and therefore tests whether adaptive conjecture weighting contributes beyond CVaR scoring alone.

We use the finite-family likelihood update as the conceptual posterior update. In the experiments,
this update is implemented with standard numerical stabilization to avoid posterior collapse under
noisy local observations; Appendix~\ref{app:implementation_scope} gives the details. 

\subsection{Conjectural Scenario Generation}

Let $s_t$ denote the full current moving-obstacle state required by a conjectural predictor, such as
obstacle positions together with tracked velocities or other hidden motion variables used to
propagate obstacles forward. This state is not necessarily identical to the current observation
$o_t$: the robot may observe only part of the obstacle state, or may be uncertain about velocity,
intent, or recently occluded motion. After assimilating the local observation history, each
conjecture $\theta$ maintains a belief distribution $b_t^\theta$ over the full current moving-obstacle
state $s_t$; equivalently, $b_t^\theta(s)$ denotes the belief mass or density of $s_t=s$ under
conjecture $\theta$ given $(\mathcal{H}_t,M_t)$. If the current obstacle state is fully observed,
then $b_t^\theta$ reduces to a point mass at the observed state; otherwise, it represents a
distribution over plausible hidden completions consistent with the observations.

At each control step, for scenario count $N$ and horizon $H$, \method{} samples short-horizon
futures from the posterior-weighted conjectural mixture. Equivalently, for each scenario
$i=1,\ldots,N$, it draws
\(
    \theta^{(i)}\sim q_t
\)
and
\(
    s_t^{(i)}\sim b_t^{\theta^{(i)}},
\)
and then rolls the moving obstacles forward under the sampled conjecture:
\(
    \boldsymbol{\zeta}_{t}^{(i)}
    =
    \bigl(\zeta_{t+1}^{(i)},\ldots,\zeta_{t+H}^{(i)}\bigr)
    \sim
    Q_{\theta^{(i)}}(\cdot\mid s_t^{(i)},M_t).
\)
Here $\boldsymbol{\zeta}_t^{(i)}$ denotes the entire future obstacle rollout sampled at decision
time $t$, while $\zeta_{t+k}^{(i)}$ denotes the obstacle state at future step $k$. Thus scenario
generation propagates three sources of uncertainty: uncertainty over the local motion conjecture
$\theta$, uncertainty over the full current obstacle state $s_t$, and stochasticity or model
randomness in the future obstacle rollout. The sampled futures are not used to certify the
conjectural model as correct; they are used to test candidate commands against plausible
short-horizon interactions before the robot physically commits to one of them.

\subsection{CVaR Scenario Planning}

For each candidate velocity command $u\in\mathcal{U}$, RCSP rolls out the robot under every
sampled future trajectory $\boldsymbol{\zeta}_t^{(i)}$. Let $R^{(i)}(u)$ denote progress toward the
goal and let $G^{(i)}(u)$ denote trajectory-level risk, such as inverse clearance, maximum safety
violation, or near-miss cost over the horizon. Rather than optimizing only the mean rollout reward
or the single worst sampled future, RCSP uses a tail-risk objective
\(
    J_t(u)
    =
    \frac{1}{N}\sum_{i=1}^{N} R^{(i)}(u)
    -
    \lambda\,
    \widehat{\cvar}_{\alpha}
    \bigl(\{G^{(i)}(u)\}_{i=1}^{N}\bigr).
\)
Here $\widehat{\cvar}_{\alpha}$ averages the largest-risk $\alpha$ fraction of sampled outcomes,
with $\alpha$ interpreted as the tail fraction. The planner-selected nominal command is
\(
    u^{\nom}_t\in\arg\max_{u\in\mathcal{U}}J_t(u).
\)
This objective is designed for near-miss commitment: it penalizes candidate commands whose average
rollout may look acceptable but whose high-risk tail contains closing-passage or collision-prone
futures.

\subsection{Fixed Safety Execution Layer}

The scenario planner selects a planner-selected nominal command $u^{\nom}_t$, i.e., the command
before local safety filtering. This command is passed through the fixed local safety filter in
Figure~\ref{fig:method-flowchart} before execution. The filter is CBF-style but discrete: rather
than solving a continuous quadratic program, it evaluates the finite candidate set
\(
    \mathcal{A}_t=\{u^{\nom}_t\}\cup\mathcal{U},
\)
where $\mathcal{U}$ is the same finite differential-drive velocity lattice used by the planner.
For each $u\in\mathcal{A}_t$, the filter performs a short local rollout using the current observation
$o_t$ and tracked obstacle velocities. Let $c_t$ be the current local clearance and $c_{\min}(u)$
the minimum predicted clearance along this rollout. With hard-clearance margin
$c_{\mathrm{hard}}$, a candidate is locally feasible if
\[
    c_{\min}(u)\ge c_{\mathrm{hard}}
    \quad\text{and}\quad
    \bigl(c_{\min}(u)-c_{\mathrm{hard}}\bigr)
    +\kappa\bigl(c_t-c_{\mathrm{hard}}\bigr)\ge 0,
\]
where $\kappa>0$ is a fixed barrier-style gain. The first condition enforces a clearance margin; the
second discourages commands that rapidly erode the current safety margin.
Among the finite candidates, the filter selects a command that balances short-horizon progress,
clearance, and closeness to $u^{\nom}_t$, with infeasible candidates penalized. We write the
executed command as
\(
    u^{\safe}_t=\Pi_{\safe}(u^{\nom}_t,o_t).
\)
This layer uses only current local geometry and short-horizon obstacle extrapolation: it does not
sample conjectural futures, update the posterior, evaluate CVaR, or choose a new route. The same
fixed filter is used for the shielded RCSP variants, including full posterior RCSP, fixed-predictor
RCSP, and mean-risk + filter. The CVaR-only row disables this filter. Thus comparisons among
shielded rows isolate predictive scenario-planning choices under a common execution layer.
Algorithm~\ref{alg:rcsp} summarizes the whole framework.

\begin{algorithm}[h]
\caption{Risk-sensitive conjectural scenario planning}
\label{alg:rcsp}
\small
\begin{enumerate}[leftmargin=1.4em,itemsep=0.08em,topsep=0.2em]
    \item \textbf{Require:} finite model family $\Theta$, prior distribution
    $q_0\in\Delta(\Theta)$ over conjectures, horizon $H$, scenario count $N$,
    CVaR tail fraction $\alpha$, progress-risk weight $\lambda$, and finite command set
    $\mathcal{U}$.

    \item \textbf{Observe and remember:} receive local observation $o_t$, append it to
    $\mathcal{H}_t$, update the local map memory $M_t$, and maintain obstacle-state beliefs
    $\{b_t^\theta\}_{\theta\in\Theta}$ under each conjecture.

    \item \textbf{Conjecture update:} for each $\theta\in\Theta$, compute the one-step predictive
    likelihood
    \(
        \ell_t(\theta)=L_\theta(o_t\mid\mathcal{H}_{t-1},u_{t-1}),
    \)
    which measures how well conjecture $\theta$ explains the newly observed local scene $o_t$.
    Conceptually, RCSP performs the finite-family likelihood update
    \(
        \bar q_t(\theta)=
        \frac{q_{t-1}(\theta)\ell_t(\theta)}
        {\sum_{\theta'\in\Theta}q_{t-1}(\theta')\ell_t(\theta')}.
    \)
    In the experiments, the posterior used for scenario sampling is a stabilized version of this
    finite-family update, as described in Appendix~\ref{app:implementation_scope}.

    \item \textbf{Scenario evaluation:} for each candidate velocity command
    $u\in\mathcal{U}$ and each sample $i=1,\ldots,N$:
    \begin{enumerate}[leftmargin=1.4em,itemsep=0.04em,topsep=0.05em,label*=\arabic*.]
        \item sample a conjecture $\theta^{(i)}\sim q_t$, a current moving-obstacle state
        $s_t^{(i)}\sim b_t^{\theta^{(i)}}$, and a future moving-obstacle trajectory
        \(
            \boldsymbol{\zeta}_{t}^{(i)}
            =
            \bigl(\zeta_{t+1}^{(i)},\ldots,\zeta_{t+H}^{(i)}\bigr)
            \sim
            Q_{\theta^{(i)}}(\cdot\mid s_t^{(i)},M_t),
        \)
        where $\boldsymbol{\zeta}_{t}^{(i)}$ denotes the whole future obstacle rollout sampled at
        decision time $t$, and $\zeta_{t+k}^{(i)}$ denotes its obstacle state at future step $k$;

        \item roll out the robot from its current state $x_t^r$ for $H$ steps under candidate
        command $u$ against this sampled obstacle trajectory, yielding simulated robot states
        \(
            \bigl(x_{t+1}^{r,(i)}(u),\ldots,x_{t+H}^{r,(i)}(u)\bigr);
        \)

        \item compute progress reward $R^{(i)}(u)$ and trajectory risk
        \(
            G^{(i)}(u)
            =
            \max_{1\le k\le H}
            g_{\rm risk}
            \bigl(x_{t+k}^{r,(i)}(u),\zeta_{t+k}^{(i)}\bigr),
        \)
        where larger $G^{(i)}(u)$ indicates lower predicted clearance or greater safety violation.
    \end{enumerate}

    \item \textbf{Tail-risk score:}
    \(
        J_t(u)=\frac{1}{N}\sum_{i=1}^{N}R^{(i)}(u)
        -\lambda\widehat{\cvar}_\alpha(\{G^{(i)}(u)\}_{i=1}^{N}),
    \)
    where $\widehat{\cvar}_\alpha$ averages the largest-risk $\lceil\alpha N\rceil$ sampled
    outcomes, with $\alpha$ interpreted as the tail fraction.

    \item \textbf{Act:} choose
    \(
        u_t^{\nom}\in\arg\max_{u\in\mathcal{U}}J_t(u),
    \)
    and execute
    \(
        u_t^{\safe}=\Pi_{\safe}(u_t^{\nom},o_t)
    \)
    through the fixed local safety filter.
\end{enumerate}
\end{algorithm}

\section{Experiments}

The experiments address three questions. \textbf{Q1:} Does full posterior \method{} prevent
dynamic near-miss failures in controlled simulation? \textbf{Q2:} Which components are responsible
for the gain? \textbf{Q3:} What happens on public benchmarks where mature local planners are already
strong?

\subsection{Evaluation Scope and Simulation Protocol}

We use full posterior \method{} as the canonical controller: online conjecture-posterior updating,
posterior-mixture scenario sampling, CVaR tail-risk scoring, and a fixed local execution filter.
We also evaluate controlled variants. Fixed-predictor \method{} disables posterior updating while
retaining CVaR scoring and the fixed filter. The ROS2/Gazebo Nav2 experiment is a deployment-wrapper
test in which the fixed \method{} execution filter supervises official Nav2 commands.

We evaluate on a controlled-to-public benchmark ladder:
\begin{itemize}[leftmargin=1.2em,itemsep=0.15em]
    \item \textbf{MuJoCo controlled bottlenecks:} differential-drive dynamics, local range
    observations, latency, velocity response, actuation noise, and dynamic obstacles.
    \item \textbf{ROS2/Gazebo dynamic near-miss:} TurtleBot/Nav2-style moving-obstacle settings
    with official Nav2 controller baselines.
    \item \textbf{Official DynaBARN/Jackal:} public dynamic-navigation transfer with ROS Noetic
    Jackal, DWA, and TEB \citep{nair2022dynabarn}.
\end{itemize}

\paragraph{Baseline fairness and matching.}
Across matched comparisons, all methods use the same start--goal pairs, random seeds, robot model,
obstacle schedule, and local observation interface within each environment. ROS2/Nav2 and official
DynaBARN baselines use maintained benchmark or stack implementations where available; MuJoCo
DWA/TEB/ORCA rows are style-matched simulator comparators using the same differential-drive action
interface. Hyperparameters are fixed before final evaluation and are not tuned per seed.

\begin{figure}[h]
\centering
\includegraphics[width=\textwidth]{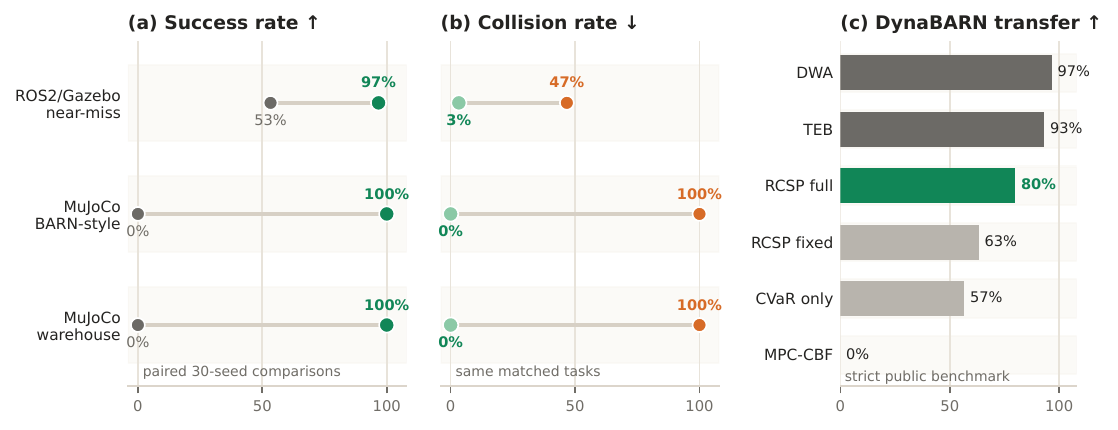}
\caption{Compact cross-benchmark summary of selected headline comparisons. The
ROS2/Gazebo panel compares Nav2 MPPI with the Nav2+RCSP execution-filter wrapper. The
MuJoCo panels compare the DWA-style local planner with full posterior RCSP on matched
controlled bottleneck tasks. The DynaBARN panel reports official benchmark success for
DWA/TEB and RCSP variants, highlighting the public-transfer boundary: maintained DWA/TEB
remain stronger on strict success, while full posterior RCSP is strongest among the
RCSP variants.}
\label{fig:core-results}
\end{figure}

Figure~\ref{fig:core-results} gives a compact overview of the selected headline
comparisons across this evaluation ladder; the detailed MuJoCo and DynaBARN metrics are
reported in Tables~\ref{tab:mujoco-controlled} and ~\ref{tab:dynabarn-ablation}.

\subsection{Main Dynamic Navigation Results}

Table~\ref{tab:mujoco-controlled} gives the final controlled MuJoCo all-variant rerun. Full
posterior \method{} and fixed-predictor \method{} both solve the medium bottleneck tasks on
success and collision. The posterior version gives better point estimates on score, safety cost,
signed minimum clearance, and SPL, at the cost of higher latency.

\begin{table}[h]
\centering
\caption{Controlled MuJoCo dynamic-bottleneck results over two environments and
30 matched seeds per environment. Full posterior RCSP preserves the strong
success/collision result and gives higher secondary point estimates than
fixed-predictor RCSP, but is slower. ``Score'' is Success $-$ Collision
$-$ 0.10 Timeout $-$ 0.03 SafetyCost; ``Min clear.'' is signed clearance,
with negative values indicating collision-margin violation.}
\label{tab:mujoco-controlled}
\footnotesize
\setlength{\tabcolsep}{3.0pt}
\resizebox{\textwidth}{!}{
\begin{tabular}{lrrrrrrrr}
\toprule
Method & Score & Success & Collision & Timeout & Safety cost & Min clear. & SPL & Lat. ms \\
\midrule
Full posterior \method{} & \textbf{0.478} & \textbf{1.000} & \textbf{0.000} & \textbf{0.000} & \textbf{17.394} & \textbf{0.135} & \textbf{1.000} & 535.669 \\
Fixed-predictor \method{} & 0.466 & \textbf{1.000} & \textbf{0.000} & \textbf{0.000} & 17.814 & 0.128 & \textbf{1.000} & 428.796 \\
Mean-risk + filter & 0.404 & 0.983 & 0.017 & \textbf{0.000} & 18.752 & 0.125 & 0.982 & 428.216 \\
CVaR-only & -1.081 & 0.183 & 0.467 & 0.350 & 25.410 & 0.069 & 0.135 & 407.711 \\
MPC-CBF & -2.205 & 0.033 & 0.917 & 0.050 & 43.894 & -0.022 & 0.032 & \textbf{2.604} \\
DWA-style & -1.460 & 0.000 & 1.000 & \textbf{0.000} & 15.325 & -0.060 & 0.000 & 20.909 \\
TEB-style & -2.997 & 0.000 & 0.983 & 0.017 & 67.051 & -0.025 & 0.000 & 0.104 \\
ORCA-style & -3.329 & 0.000 & 0.967 & 0.033 & 78.645 & -0.060 & 0.000 & 9.437 \\
\bottomrule
\end{tabular}
}
\end{table}

In the ROS2/Gazebo Nav2 comparison, the evaluated variant is a Nav2+\method{} execution filter:
an official Nav2 controller provides the nominal command, and the fixed \method{} execution filter
supervises the velocity command. This tests whether the local execution filter can reduce dynamic
near-miss failures inside an otherwise standard navigation stack; it does not isolate the full
posterior scenario planner. Against the strongest official Nav2 baseline, MPPI, this wrapper achieves
0.967 success and 0.033 collision, compared with 0.533 success and 0.467 collision for MPPI.
Safety cost and collision are related but distinct: safety cost accumulates conservative slow-motion
and proximity exposure over successful episodes, whereas collision is a terminal safety failure.

\subsection{Ablations and Public-Benchmark Boundary}

We use official DynaBARN/Jackal as the main external transfer test because it is not tailored to
our controlled near-miss environments and includes mature benchmark navigation stacks. This
setting is therefore used both as an ablation test for \method{} and as a boundary case for the
paper's claim. Table~\ref{tab:dynabarn-ablation} evaluates full posterior \method{} against its
fixed-predictor and lower-level ablations, as well as maintained DWA and TEB baselines.

\begin{table}[h]
\centering
\caption{Official DynaBARN/Jackal transfer. DWA/TEB remain stronger under the official move-base
success signal. The full posterior/expanded-family \method{} variant gives better point estimates
than fixed-predictor \method{} and lower-level ablations on endpoint and adapter safety diagnostics,
but the full-vs-fixed success advantage is directional rather than conclusive.}
\label{tab:dynabarn-ablation}
\footnotesize
\setlength{\tabcolsep}{3.2pt}
\renewcommand{\arraystretch}{1.08}
\resizebox{\textwidth}{!}{
\begin{tabular}{lrrrrr}
\toprule
Method & Official success & Final-pose comp. & Adapter collision & Min clear. & Final dist. \\
\midrule
DWA & \textbf{0.967} & 0.300 & n/a & n/a & 1.281 \\
TEB & 0.933 & 0.267 & n/a & n/a & 1.535 \\
Fixed-predictor \method{} & 0.633 & 0.967 & 0.367 & 0.930 & 0.523 \\
Full posterior \method{} & 0.800 & \textbf{1.000} & \textbf{0.200} & \textbf{1.026} & \textbf{0.300} \\
CVaR-only & 0.567 & 0.733 & 0.333 & 0.904 & 1.100 \\
MPC-CBF & 0.000 & 0.000 & 0.367 & 0.471 & 4.620 \\
\bottomrule
\end{tabular}
}

\vspace{2pt}
\begin{minipage}{0.98\linewidth}
\footnotesize
\emph{Notes.} ``Official success'' is the benchmark move-base status. ``Final-pose comp.'' uses a
0.45 m final-pose threshold and is not the same as official success. ``Adapter collision'' is the
collision label from the \method{} adapter's signed-clearance monitor; it is not a generic near-miss
threshold. ``Min clear.'' is signed clearance. Adapter-collision and min-clearance labels are n/a for
DWA/TEB because official logs do not expose this monitor.
\end{minipage}
\end{table}

The results support a targeted mechanism claim rather than a universal navigation-stack
claim. In the controlled medium bottleneck tasks, full posterior RCSP achieves strong
success and collision-avoidance performance. The ablations indicate that CVaR scoring
and the fixed execution filter are central to the controlled-task gains. Restoring the
online conjecture posterior gives higher secondary point estimates than the
fixed-predictor variant, but does not improve success or collision in the medium table
because both shielded variants already solve those tasks.

On DynaBARN/Jackal, the full posterior/expanded-family RCSP variant gives better point
estimates than fixed-predictor RCSP, CVaR-only, and MPC-CBF on endpoint and adapter
safety diagnostics. This evidence is directional rather than conclusive: the paired
official-success advantage over fixed-predictor RCSP has a bootstrap interval crossing
zero, and posterior diagnostics are diffuse in this interface.

\section{Discussion and Limitations}

The experiments support a targeted mechanism claim rather than universal navigation SOTA.
\method{} is strongest when dynamic obstacles create predictive near-miss commitments, where
short-horizon local clearance is misleading. It is less compelling when performance is dominated
by mature path tracking, robot-interface details, or fully observed local constraints: official
DynaBARN shows this boundary, with DWA/TEB achieving higher strict success while full
\method{} has better point estimates than its ablations. The main limitation is that the study is
simulation-only; it does not validate real sensing noise, physical latency, calibration, actuation
delays, wheel--ground interaction, or long-horizon autonomy on hardware. \method{} also incurs
higher latency and less smooth control than mature local controllers, motivating faster scenario
evaluation, integration with stronger navigation stacks, and physical-robot validation.

\section{Conclusion}

RCSP studies a specific failure mode in dynamic navigation: locally safe actions can
commit a robot to future near misses before low-level safety filters can recover. In
controlled dynamic near-miss and bottleneck simulations, full posterior RCSP combines
posterior-mixture scenario sampling, CVaR tail-risk scoring, and fixed local execution
filtering to achieve strong success and collision-avoidance performance. The posterior
variant gives higher secondary point estimates than fixed-predictor RCSP at higher
latency, but does not improve success/collision in the medium controlled table because
both shielded variants already solve those tasks. ROS2/Gazebo evaluates a Nav2+RCSP
execution-filter wrapper rather than the full posterior planner. Public DynaBARN
transfer shows both promise and boundary: full posterior RCSP gives better point
estimates than its own ablations, but mature DWA/TEB controllers remain stronger on
official strict success. The method is therefore best viewed as a simulation-tested
predictive-risk layer to be integrated with, rather than replace, robust navigation
stacks.

\bibliography{references}

\clearpage
\appendix
\input{appendix}

\end{document}

%% file: appendix.tex
\section*{Appendix}

\section{Implementation Scope}
\label{app:implementation_scope}

The canonical full posterior RCSP controller consists of posterior conjecture updating,
posterior-mixture scenario sampling, CVaR tail-risk scoring, and a fixed local execution
filter. Some experiments intentionally instantiate controlled variants: fixed-predictor
rows disable posterior updating, CVaR-only disables the fixed execution filter,
ROS2/Nav2 rows evaluate the fixed execution filter as a supervisor over official Nav2
commands, and full posterior rows use the posterior-mixture controller.

Environment-specific adapters translate observations, simulator states, and velocity
commands into the relevant robot interface. They should not be read as identical
controller deployments across all environments.

\paragraph{Stabilized posterior weights.}
Here $q_t$ denotes the posterior weight over conjectural models, not the
obstacle-state belief. For each conjecture $\theta$, the corresponding
obstacle-state belief $b_t^\theta$ and rollout model $Q_\theta$ are used to
generate future scenarios; $q_t$ determines how much probability mass is
assigned to each conjecture when sampling those scenarios.

Algorithm~\ref{alg:rcsp} presents the conceptual finite-family likelihood update
$q_t(\theta)\propto q_{t-1}(\theta)\ell_t(\theta)$. In code, we use standard numerical
stabilizers for this finite-family filter. In particular, likelihood tempering replaces the
likelihood factor by
\[
    \ell_t(\theta)^{1/\tau},
\]
so the tempered update has the form
\[
    q_t(\theta)
    \propto
    q_{t-1}(\theta)\ell_t(\theta)^{1/\tau},
\]
where $\tau>0$ is a temperature. Smaller $\tau$ sharpens the update, while larger $\tau$
keeps the posterior more diffuse. Equivalently, in log space, this corresponds to adding
$\log \ell_t(\theta)/\tau$ to the posterior log weight.

After normalization, we apply a small probability floor so that every conjecture retains
nonzero support. In expanded-family runs, finite scenario samples may be drawn from the
top-$K$ posterior mixture after renormalization; this affects scenario allocation only and
is not hard MAP model selection. These stabilizers prevent premature posterior collapse
under noisy local obstacle tracks, but they do not change the controller's decision
structure: \method{} still samples futures from a posterior mixture, scores candidate
commands by CVaR tail risk, and executes the selected command through the fixed local
safety layer.

\section{Information-State and Conjectural Model Formulation}
\label{app:information_state}

This section gives the information-state view behind \method{} and aligns the notation with
Algorithm~\ref{alg:rcsp}. Let $\mathcal{H}_t$ denote the history of observations and executed
commands before choosing the next command:
\[
    \mathcal{H}_t =
    \{(o_\tau,u_\tau)\}_{\tau=0}^{t-1}\cup\{o_t\}.
\]
All admissible decisions are measurable with respect to the information generated by
$\mathcal{H}_t$. \method{} summarizes this history by
\[
    I_t=(M_t,\{b_t^\theta\}_{\theta\in\Theta},q_t,g).
\]
Here $M_t$ is the accumulated local map or occupancy memory, $b_t^\theta$ is the belief over nearby
moving-obstacle states under conjecture $\theta$, $q_t\in\Delta(\Theta)$ is the posterior weight
over conjectures, and $g$ is the goal.

\paragraph{Knowledge state.}
Let $V(x^r_t)$ denote the region visible to the robot from pose $x^r_t$. If $E_t$ is the explored
region and $\xi^{\mathrm{stat}}$ is the static environment, the map memory can be written as
\[
    M_t=(E_t,\xi^{\mathrm{stat}}|_{E_t}),
    \qquad
    E_{t+1}=E_t\cup V(x^r_{t+1}).
\]
Thus $M_t$ captures accumulated static information: once part of the local environment has been
observed, it remains available to the controller.

\paragraph{Belief under a conjectured dynamics model.}
Let $s_t$ denote the full current state of nearby moving obstacles, including positions, tracked
velocities, or other hidden motion variables needed by a conjectural predictor. Given conjecture
$\theta$, the scenario model $Q_\theta$ induces an obstacle-transition kernel
$Q^{\mathrm{obs}}_\theta$. The belief update follows the usual prediction--correction form. First,
the controller predicts the next obstacle-state belief by marginalizing over the current hidden state:
\[
    \tilde b_{t+1}^{\theta}(s')
    =
    \sum_s
    b_t^\theta(s)
    Q^{\mathrm{obs}}_{\theta}(s'\mid s,x^r_t,u_t,M_t).
\]
Here $b_t^\theta(s)$ is the current belief that the obstacle state is $s$, and
$Q^{\mathrm{obs}}_{\theta}(s'\mid s,x^r_t,u_t,M_t)$ is the conjectured probability of moving from
$s$ to $s'$. The product gives the contribution of previous state $s$ to future state $s'$, and the
sum marginalizes over all possible previous states. Then the new observation reweights the
prediction:
\[
    b_{t+1}^{\theta}(s')
    \propto
    \tilde b_{t+1}^{\theta}(s')L_{\theta}(o_{t+1}\mid s',M_{t+1}).
\]
The likelihood $L_\theta$ measures how compatible the new local observation is with candidate
state $s'$. For continuous obstacle states, the sum is replaced by an integral.

\paragraph{Conjecture update and predictive KL interpretation.}
The conjectural model weights are updated by predictive performance. Let
$m_\theta(o_{t+1}\mid\mathcal{H}_t,u_t)$ denote the one-step predictive observation distribution
induced by the conjectured model $Q_\theta$. The cumulative predictive log-likelihood of conjecture
$\theta$ is
\[
    \sum_{\tau=0}^{t-1}
    \log m_\theta(o_{\tau+1}\mid\mathcal{H}_\tau,u_\tau).
\]
When the model class is misspecified, high posterior weight should not be interpreted as recovering
the true environment model. Let $\mu_u$ denote the true observation distribution induced by applying
the command or local policy $u$ in the environment. In a stationary local regime, likelihood-based
updating favors conjectures whose predictive observation distributions are closest to $\mu_u$ in
predictive Kullback--Leibler divergence:
\[
    \theta^\star(u)
    \in
    \arg\min_{\theta\in\Theta}
    D_{\mathrm{KL}}
    \bigl(\mu_u\,\|\,m_\theta(\cdot\mid u)\bigr).
\]
This is the misspecified-model interpretation behind the conjecture update: a model can be
predictively useful even when it is not a literal true model of the world. Appendix~\ref{app:berk_selection}
gives the corresponding finite-family likelihood-selection statement.

\section{Model-Conditional Theory for Conjectural CVaR Planning}
\label{app:model_conditional_theory}

This appendix gives a model-conditional interpretation of \method{}. The results below do not
constitute a closed-loop real-world safety guarantee. Instead, they clarify what is supported by the
algorithmic structure when the conjectured scenario model is treated as the decision model:
likelihood-based updates select predictive conjectures under misspecification, finite scenario
samples approximate model-implied CVaR tail risk, and an idealized model--decision--data fixed
point exists under standard compactness and continuity assumptions.

\subsection{Berk-Style Selection of Conjectural Predictors}
\label{app:berk_selection}
The analysis below uses the idealized untempered finite-family likelihood update.
The implementation uses the stabilized update described in
Appendix~\ref{app:implementation_scope}; these numerical regularizers are used for
finite-family filtering under noisy local observations and are not modeled in the
idealized statements below.

This subsection formalizes the conjecture-update step used by \method{}. We call the update
\emph{Berk-style} because, under model misspecification and suitable regularity conditions,
likelihood-based posterior updating need not recover a literal true model. Instead, posterior mass
concentrates on models that are closest to the data-generating process in predictive
Kullback--Leibler divergence \citep{berk1966limiting}. This perspective is also related to
Berk--Nash equilibrium, where an agent's decision determines the data distribution that subsequently
selects the predictive model \citep{esponda2016berk}. In our setting, this corresponds to a
model--decision--data loop: a conjecture affects the planned command, the command affects future
observations, and those observations update the conjecture.

In the main text, the posterior weight over conjectures is denoted by
$q_t\in\Delta(\Theta)$. In this subsection we write the same object as
$w_t\in\Delta(\Theta)$ to avoid confusion with predictive observation distributions. The model
family $\Theta$ is finite, and each $\theta\in\Theta$ represents one local hypothesis about how
nearby moving obstacles may evolve and how future observations may be generated.

Given the current history $\mathcal{H}_t$ and the command executed at time $t$, conjecture
$\theta$ induces a one-step predictive distribution over the next local observation. We denote its
likelihood or density by
\[
    m_\theta(o_{t+1}\mid \mathcal{H}_t,u_t)>0.
\]
Here $o_{t+1}$ is the same local observation variable used in the main text. The quantity
$m_\theta(o_{t+1}\mid \mathcal{H}_t,u_t)$ should be read as: ``how likely was the newly observed
local scene under conjecture $\theta$?'' It is induced by the conjectured scenario model
$Q_\theta$ after marginalizing over latent moving-obstacle states and other unobserved quantities.
The strict positivity assumption avoids degenerate likelihood ratios; in implementation this
corresponds to assigning nonzero probability or density to observations that may occur.

The posterior update is the usual finite-family likelihood update:
\[
    w_{t+1}(\theta)
    =
    \frac{
    w_t(\theta)\,
    m_\theta(o_{t+1}\mid \mathcal{H}_t,u_t)}
    {\sum_{\theta'\in\Theta}
    w_t(\theta')\,
    m_{\theta'}(o_{t+1}\mid \mathcal{H}_t,u_t)}.
\]
Thus, a conjecture receives more weight when it assigns higher predictive likelihood to the
observation that actually occurs. Conversely, a conjecture whose predictions are repeatedly
inconsistent with the observed local interaction history loses weight.

Unrolling this recursion over $T$ observations gives
\[
    w_T(\theta)
    =
    \frac{
    w_0(\theta)
    \exp\left(\sum_{t=0}^{T-1}
    \log m_\theta(o_{t+1}\mid \mathcal{H}_t,u_t)\right)}
    {\sum_{\theta'\in\Theta}
    w_0(\theta')
    \exp\left(\sum_{t=0}^{T-1}
    \log m_{\theta'}(o_{t+1}\mid \mathcal{H}_t,u_t)\right)}.
\]
This expression shows that the posterior weight is controlled by cumulative predictive
log-likelihood. Under misspecification, the update therefore selects conjectures that are most
predictive of the observations generated by the closed-loop robot, rather than necessarily selecting
a literal true model.

\subsection{Why CVaR Rather Than the Single Worst Rollout}
\label{app:cvar_vs_worst_case}

A natural conservative alternative to the CVaR objective is the single-worst-rollout score
\[
    \widehat J^{\mathrm{worst}}_{t,N}(u)
    =
    \frac{1}{N}\sum_{i=1}^{N}R^{(i)}(u)
    -
    \lambda \max_{1\le i\le N}G^{(i)}(u).
\]
This objective is easy to justify from a safety perspective, but it is brittle under finite scenario
sampling. If one sampled future is an extreme low-probability or noisy rollout, the entire command
score is dominated by that single sample. In dynamic navigation, this can make the robot overly
conservative: it may brake, hesitate, take unnecessarily long detours, or fail to complete the route
even when most plausible futures remain safe.

The mean-risk objective has the opposite failure mode:
\[
    \widehat J^{\mathrm{mean}}_{t,N}(u)
    =
    \frac{1}{N}\sum_{i=1}^{N}R^{(i)}(u)
    -
    \lambda \frac{1}{N}\sum_{i=1}^{N}G^{(i)}(u).
\]
It can average away rare but dangerous futures and is therefore too optimistic for near-miss
commitment. Empirical CVaR provides an intermediate risk measure:
\[
    \widehat J^{\mathrm{CVaR}}_{t,N}(u)
    =
    \frac{1}{N}\sum_{i=1}^{N}R^{(i)}(u)
    -
    \lambda \widehat{\mathrm{CVaR}}_{\alpha}
    \bigl(\{G^{(i)}(u)\}_{i=1}^{N}\bigr),
\]
where $\widehat{\mathrm{CVaR}}_{\alpha}$ averages the largest-risk
$\lceil \alpha N\rceil$ samples. When $\alpha=1$, the risk term reduces to the empirical mean risk;
when $\lceil \alpha N\rceil=1$, it reduces to the empirical worst-case risk. Intermediate values of
$\alpha$ retain sensitivity to dangerous tail futures while reducing dependence on a single extreme
sample.

This distinction is why the main implementation uses CVaR rather than a hard worst-case score. The
theory in Appendix~\ref{app:cvar_approximation} therefore analyzes model-conditional approximation
of the CVaR objective rather than chance-feasibility guarantees for hard sampled constraints.

\subsection{Model-Conditional Empirical CVaR Approximation}
\label{app:cvar_approximation}

This subsection explains why a finite number of sampled futures can be used to estimate the
tail-risk objective optimized by \method{}. The statement is model-conditional: it treats the current
posterior-weighted scenario model as the decision model and does not claim safety under the true
environment when that model is misspecified.

Recall that $w_t\in\Delta(\Theta)$ denotes the posterior weight over conjectures. At information
state $I_t$, the posterior-weighted scenario model is
\[
    Q_{w_t}(\cdot\mid I_t)
    =
    \sum_{\theta\in\Theta}
    w_t(\theta)Q_\theta(\cdot\mid I_t).
\]
For each candidate command $u\in\mathcal{U}$, the planner evaluates $u$ under sampled futures
\[
    \boldsymbol{\zeta}_t^{(i)}\sim Q_{w_t}(\cdot\mid I_t),
    \qquad i=1,\ldots,N.
\]
These sampled futures describe plausible moving-obstacle evolutions under the current information
state. The command $u$ affects the simulated robot rollout inside each sampled future, written as
$x_{t+k}^{r,(i)}(u)$, and therefore affects the reward and risk values computed from that
rollout. Let $R(u,\zeta)$ denote progress reward and let $G(u,\zeta)$ denote trajectory-level risk,
where larger $G$ means lower clearance or greater safety violation. We assume bounded reward and
bounded risk,
\[
    G(u,\zeta)\in[0,B_G],
    \qquad
    R(u,\zeta)\in[R_{\min},R_{\max}],
\]
and define the reward range $B_R=R_{\max}-R_{\min}$. Here $B_G$ denotes a finite upper bound on
the risk values under the finite-horizon scenario model; the implementation may use collision
penalties or scaled safety-margin violations, so the bound is not assumed to be one.

We next define the tail-risk functional used in the planner. For a risk random variable
$X\in[0,B_G]$, the cumulative distribution function
\[
    F_X(x)=\Pr(X\le x)
\]
gives the fraction of outcomes whose risk is at most $x$. The quantile function reverses this
question:
\[
    F_X^{-1}(p)
    =
    \inf\{x\in[0,B_G]:F_X(x)\ge p\},
    \qquad p\in[0,1].
\]
It is the smallest risk threshold $x$ such that at least a $p$ fraction of outcomes have risk no
larger than $x$. Thus, the $(1-\alpha)$ quantile is the threshold where the worst $\alpha$ tail
begins. We define the upper-tail CVaR over this worst $\alpha$ fraction as
\[
    \rho_\alpha(X)
    =
    \frac{1}{\alpha}
    \int_{1-\alpha}^{1}F_X^{-1}(p)\,dp,
    \qquad \alpha\in(0,1].
\]
Thus, $\alpha$ denotes the \emph{tail fraction}, not the quantile level. For example,
$\alpha=0.1$ means that CVaR averages the worst $10\%$ of outcomes, which lie above the
$0.9$ quantile. Equivalently, the same tail average can be written in the standard optimization form
\[
    \rho_\alpha(X)
    =
    \inf_{\eta\in[0,B_G]}
    \left\{
        \eta+\frac{1}{\alpha}\mathbb{E}\bigl[(X-\eta)_+\bigr]
    \right\}.
\]
This form has a simple threshold interpretation. The scalar $\eta$ is a candidate risk threshold,
and $(X-\eta)_+=\max\{X-\eta,0\}$ counts only the excess risk above that threshold. If $\eta$ is
chosen too low, many outcomes exceed it and the excess term is large. If $\eta$ is chosen too high,
the threshold term $\eta$ itself is large. The minimizing threshold therefore balances these two
effects. For a continuous distribution, differentiating the objective with respect to $\eta$ gives the
condition
\[
    \Pr(X>\eta)=\alpha,
\]
so the minimizer is the $(1-\alpha)$-quantile, i.e., the point where the worst $\alpha$ tail begins.
At this threshold, the objective equals the average risk in that worst tail. When the distribution has
atoms, the minimizing threshold may not be unique; the quantile-integral definition above gives the
corresponding fractional-tail average.

For each command $u$, define the population model-conditional tail risk
\[
    \rho_{\alpha,t}(u)
    :=
    \rho_\alpha\bigl(G(u,\zeta)\bigr),
    \qquad
    \zeta\sim Q_{w_t}(\cdot\mid I_t).
\]
The empirical version $\widehat\rho_{\alpha,N,t}(u)$ is computed from the sampled risks
\[
    \{G(u,\zeta_t^{(i)})\}_{i=1}^{N}.
\]
In implementation, this means sorting the sampled risks and averaging the largest-risk
$\lceil\alpha N\rceil$ outcomes, up to the usual rounding or fractional-weight convention for the
finite tail size. The notation $\widehat\rho_{\alpha,N,t}(u)$ is the theoretical counterpart of the empirical
$\widehat{\cvar}_\alpha(\{G^{(i)}(u)\}_{i=1}^N)$ used in Algorithm~\ref{alg:rcsp}.

\paragraph{Proposition C.2 (uniform empirical CVaR approximation).}
For finite $\mathcal{U}$ and bounded risks $G(u,\zeta)\in[0,B_G]$, with probability at least
$1-\delta$ over the scenario samples,
\[
    \sup_{u\in\mathcal{U}}
    \left|
    \widehat{\rho}_{\alpha,N,t}(u)-\rho_{\alpha,t}(u)
    \right|
    \le
    \frac{B_G}{\alpha}
    \sqrt{\frac{\log(2|\mathcal{U}|/\delta)}{2N}} .
\]

\paragraph{Proof.}
Fix a command $u\in\mathcal{U}$. Under the posterior-weighted scenario model
$\zeta\sim Q_{w_t}(\cdot\mid I_t)$, the risk $G(u,\zeta)$ is a random variable supported on
$[0,B_G]$. Let $F_u$ be its population cumulative distribution function,
\[
    F_u(x)
    =
    \Pr_{\zeta\sim Q_{w_t}(\cdot\mid I_t)}
    \bigl(G(u,\zeta)\le x\bigr),
\]
and let $\widehat F_{u,N}$ be the empirical CDF formed from the sampled risks
$\{G(u,\zeta_t^{(i)})\}_{i=1}^{N}$.

The first step is to control the sampling error of this empirical CDF. By the
Dvoretzky--Kiefer--Wolfowitz inequality, for any fixed $u$ and any $\varepsilon>0$,
\[
    \Pr\!\left(
        \sup_x|\widehat F_{u,N}(x)-F_u(x)|>\varepsilon
    \right)
    \le
    2e^{-2N\varepsilon^2}.
\]
Thus, with high probability, the empirical risk distribution is uniformly close to the
model-implied risk distribution.

The second step converts this uniform CDF error into a Wasserstein error. In one dimension,
the Wasserstein-1 distance between two distributions on $[0,B_G]$ can be written as the area
between their CDFs:
\[
    W_1(\widehat F_{u,N},F_u)
    =
    \int_0^{B_G}
    |\widehat F_{u,N}(x)-F_u(x)|\,dx .
\]
Therefore,
\[
    W_1(\widehat F_{u,N},F_u)
    \le
    B_G\sup_x|\widehat F_{u,N}(x)-F_u(x)|.
\]
Combining this with the DKW inequality gives, for fixed $u$,
\[
    \Pr\!\left(
        W_1(\widehat F_{u,N},F_u)>B_G\varepsilon
    \right)
    \le
    2e^{-2N\varepsilon^2}.
\]

The third step uses the quantile representation of upper-tail CVaR. If $F^{-1}$ and
$\widehat F^{-1}$ are the population and empirical quantile functions, then
\[
    \rho_\alpha
    =
    \frac{1}{\alpha}\int_{1-\alpha}^{1}F^{-1}(p)\,dp,
    \qquad
    \widehat \rho_\alpha
    =
    \frac{1}{\alpha}\int_{1-\alpha}^{1}\widehat F^{-1}(p)\,dp .
\]
Hence,
\[
    \left|
    \widehat{\rho}_{\alpha,N,t}(u)-\rho_{\alpha,t}(u)
    \right|
    \le
    \frac{1}{\alpha}
    \int_{1-\alpha}^{1}
    \left|\widehat F_{u,N}^{-1}(p)-F_u^{-1}(p)\right|\,dp .
\]
Since the integral over the tail is bounded by the integral over the full quantile interval,
\[
    \int_{1-\alpha}^{1}
    \left|\widehat F_{u,N}^{-1}(p)-F_u^{-1}(p)\right|\,dp
    \le
    \int_{0}^{1}
    \left|\widehat F_{u,N}^{-1}(p)-F_u^{-1}(p)\right|\,dp
    =
    W_1(\widehat F_{u,N},F_u),
\]
where the last equality is the one-dimensional quantile representation of $W_1$. Therefore,
\[
    \left|
    \widehat{\rho}_{\alpha,N,t}(u)-\rho_{\alpha,t}(u)
    \right|
    \le
    \frac{1}{\alpha}
    W_1(\widehat F_{u,N},F_u).
\]
Using the previous Wasserstein bound, for fixed $u$,
\[
    \Pr\!\left(
    \left|
    \widehat{\rho}_{\alpha,N,t}(u)-\rho_{\alpha,t}(u)
    \right|
    >
    \frac{B_G\varepsilon}{\alpha}
    \right)
    \le
    2e^{-2N\varepsilon^2}.
\]

Finally, we make the bound uniform over the finite command set. Applying a union bound over
$u\in\mathcal{U}$ gives
\[
    \Pr\!\left(
    \sup_{u\in\mathcal{U}}
    \left|
    \widehat{\rho}_{\alpha,N,t}(u)-\rho_{\alpha,t}(u)
    \right|
    >
    \frac{B_G\varepsilon}{\alpha}
    \right)
    \le
    2|\mathcal{U}|e^{-2N\varepsilon^2}.
\]
Setting
\[
    2|\mathcal{U}|e^{-2N\varepsilon^2}=\delta
\]
gives
\[
    \varepsilon
    =
    \sqrt{\frac{\log(2|\mathcal{U}|/\delta)}{2N}}.
\]
Substituting this value of $\varepsilon$ proves
\[
    \sup_{u\in\mathcal{U}}
    \left|
    \widehat{\rho}_{\alpha,N,t}(u)-\rho_{\alpha,t}(u)
    \right|
    \le
    \frac{B_G}{\alpha}
    \sqrt{\frac{\log(2|\mathcal{U}|/\delta)}{2N}}
\]
with probability at least $1-\delta$.
\hfill$\square$

\paragraph{Proposition C.3 (uniform objective approximation and near-optimality).}
With probability at least $1-\delta$,
\[
    \sup_{u\in\mathcal{U}}
    \left|
    \widehat J_{t,N}(u)-J_t(u)
    \right|
    \le
    \varepsilon_N(\delta),
\]
where
\[
    \varepsilon_N(\delta)
    =
    \left(B_R+\frac{\lambda B_G}{\alpha}\right)
    \sqrt{\frac{\log(4|\mathcal{U}|/\delta)}{2N}} .
\]
Consequently, if
\[
    \widehat u_N\in\arg\max_{u\in\mathcal{U}}\widehat J_{t,N}(u),
    \qquad
    u^\star\in\arg\max_{u\in\mathcal{U}}J_t(u),
\]
then, with the same probability,
\[
    J_t(u^\star)-J_t(\widehat u_N)
    \le
    2\varepsilon_N(\delta).
\]

\paragraph{Proof.}
The objective has two parts: an expected progress term and a CVaR risk penalty. We control their
sampling errors separately and then add the two bounds.

First consider the reward term. For each fixed command $u$, the random variables
$R(u,\zeta_t^{(1)}),\ldots,R(u,\zeta_t^{(N)})$ are bounded in
$[R_{\min},R_{\max}]$, with range $B_R=R_{\max}-R_{\min}$. By Hoeffding's inequality,
\[
    \Pr\!\left(
    \left|
    \frac{1}{N}\sum_{i=1}^{N}R(u,\zeta_t^{(i)})
    -
    \mathbb{E}_{\zeta\sim Q_{w_t}(\cdot\mid I_t)}[R(u,\zeta)]
    \right|
    >
    B_R\varepsilon
    \right)
    \le
    2e^{-2N\varepsilon^2}.
\]
Applying a union bound over the finite command set $\mathcal{U}$ gives
\[
    \Pr\!\left(
    \sup_{u\in\mathcal{U}}
    \left|
    \frac{1}{N}\sum_{i=1}^{N}R(u,\zeta_t^{(i)})
    -
    \mathbb{E}_{\zeta\sim Q_{w_t}(\cdot\mid I_t)}[R(u,\zeta)]
    \right|
    >
    B_R\varepsilon
    \right)
    \le
    2|\mathcal{U}|e^{-2N\varepsilon^2}.
\]
Setting
\[
    2|\mathcal{U}|e^{-2N\varepsilon^2}=\delta/2
\]
yields
\[
    \varepsilon
    =
    \sqrt{\frac{\log(4|\mathcal{U}|/\delta)}{2N}} .
\]
Therefore, with probability at least $1-\delta/2$,
\[
    \sup_{u\in\mathcal{U}}
    \left|
    \frac{1}{N}\sum_{i=1}^{N}R(u,\zeta_t^{(i)})
    -
    \mathbb{E}_{\zeta\sim Q_{w_t}(\cdot\mid I_t)}[R(u,\zeta)]
    \right|
    \le
    B_R
    \sqrt{\frac{\log(4|\mathcal{U}|/\delta)}{2N}} .
\]

Second, Proposition~C.2 applied with confidence level $\delta/2$ gives, with probability at least
$1-\delta/2$,
\[
    \sup_{u\in\mathcal{U}}
    \left|
    \widehat{\rho}_{\alpha,N,t}(u)-\rho_{\alpha,t}(u)
    \right|
    \le
    \frac{B_G}{\alpha}
    \sqrt{\frac{\log(4|\mathcal{U}|/\delta)}{2N}} .
\]

By another union bound, both the reward approximation event and the CVaR approximation event hold
simultaneously with probability at least $1-\delta$. Recall that by definition, the population objective and empirical objective are
\[
    J_t(u)
    =
    \mathbb{E}_{\zeta\sim Q_{w_t}(\cdot\mid I_t)}[R(u,\zeta)]
    -
    \lambda\rho_{\alpha,t}(u),
\]
and
\[
    \widehat J_{t,N}(u)
    =
    \frac{1}{N}\sum_{i=1}^{N}R(u,\zeta_t^{(i)})
    -
    \lambda\widehat\rho_{\alpha,N,t}(u).
\] So on this event, for every $u\in\mathcal U$,
\[
\begin{aligned}
    \left|
    \widehat J_{t,N}(u)-J_t(u)
    \right|
    &=
    \left|
    \left[
    \frac{1}{N}\sum_{i=1}^{N}R(u,\zeta_t^{(i)})
    -
    \mathbb{E}_{\zeta\sim Q_{w_t}(\cdot\mid I_t)}[R(u,\zeta)]
    \right]
    -
    \lambda
    \left[
    \widehat{\rho}_{\alpha,N,t}(u)-\rho_{\alpha,t}(u)
    \right]
    \right|  \\
    &\le
    \left|
    \frac{1}{N}\sum_{i=1}^{N}R(u,\zeta_t^{(i)})
    -
    \mathbb{E}_{\zeta\sim Q_{w_t}(\cdot\mid I_t)}[R(u,\zeta)]
    \right|
    +
    \lambda
    \left|
    \widehat{\rho}_{\alpha,N,t}(u)-\rho_{\alpha,t}(u)
    \right|  \\
    &\le
    \left(B_R+\frac{\lambda B_G}{\alpha}\right)
    \sqrt{\frac{\log(4|\mathcal{U}|/\delta)}{2N}} .
\end{aligned}
\]
Taking the supremum over $u\in\mathcal U$ proves the uniform objective approximation.

It remains to show the near-optimality statement. On the same event, every empirical objective value
is within $\varepsilon_N$ of its population value:
\[
    |\widehat J_{t,N}(u)-J_t(u)|\le \varepsilon_N
    \qquad\forall u\in\mathcal U.
\]
Let $u^\star$ maximize the population objective and let $\widehat u_N$ maximize the empirical
objective. Then
\[
    J_t(u^\star)
    \le
    \widehat J_{t,N}(u^\star)+\varepsilon_N
\]
because the empirical value of $u^\star$ can be at most $\varepsilon_N$ below its population value.
Since $\widehat u_N$ maximizes the empirical objective,
\[
    \widehat J_{t,N}(u^\star)
    \le
    \widehat J_{t,N}(\widehat u_N).
\]
Finally, the empirical value of $\widehat u_N$ can be at most $\varepsilon_N$ above its population
value:
\[
    \widehat J_{t,N}(\widehat u_N)
    \le
    J_t(\widehat u_N)+\varepsilon_N .
\]
Combining the three inequalities gives
\[
    J_t(u^\star)
    \le
    J_t(\widehat u_N)+2\varepsilon_N,
\]
or equivalently,
\[
    J_t(u^\star)-J_t(\widehat u_N)\le 2\varepsilon_N.
\]
\hfill$\square$

\paragraph{Interpretation.}
Propositions C.2 and C.3 justify the use of finite scenario samples for comparing candidate
commands by model-implied tail risk. They do not imply that the chosen command is safe in the true
environment. The guarantee is conditional on the posterior-weighted scenario model
$Q_{w_t}(\cdot\mid I_t)$ and on the quality of the observation and robot-interface models. If an
interactive implementation instead uses command-dependent scenario sampling
$Q_{w_t}(\cdot\mid I_t,u)$, the same finite-command-set argument applies after replacing
$Q_{w_t}(\cdot\mid I_t)$ by $Q_{w_t}(\cdot\mid I_t,u)$ throughout.

\subsection{Idealized Berk--CVaR Fixed Point}
\label{app:berk_cvar_fixed_point}

The previous subsections analyze the two ingredients of \method{} separately. The likelihood update
selects conjectures that predict the observed data well, while finite scenario samples approximate
the model-implied CVaR objective. We now give an idealized fixed-point interpretation that couples
these two ingredients. This is a structural consistency statement, not a convergence theorem for the
implemented receding-horizon controller.

To keep notation light, this subsection suppresses the current information state \(I_t\). Think of
\(u\) as a candidate local command or stationary local decision rule in a fixed local regime. For this
idealized fixed-point statement, let \(\mathcal U_{\rm id}\) denote a relaxed compact convex command
set, which is distinct from the finite command lattice used in the implementation. The
posterior-weighted scenario model is
\[
    Q_w(\cdot\mid u)
    =
    \sum_{\theta\in\Theta}w(\theta)Q_\theta(\cdot\mid u),
\]
and the corresponding model-conditional CVaR objective is
\[
    J(w,u)
    =
    \mathbb{E}_{\zeta\sim Q_w(\cdot\mid u)}[R(u,\zeta)]
    -
    \lambda \rho_\alpha\!\left(G(u,\zeta)\right).
\]
Here \(R(u,\zeta)\) is the progress reward and \(G(u,\zeta)\) is the trajectory-level risk under
scenario \(\zeta\). The CVaR planning response to conjecture \(w\) is
\[
    A(w)=\arg\max_{u\in\mathcal U_{\rm id}}J(w,u).
\]

The other direction of the loop is data-driven model selection. When command or local policy \(u\)
is applied in the true environment, it induces an observation distribution
\(\mu_u\in\Delta(\mathcal O)\) over the observation space \(\mathcal O\). Each conjecture
\(\theta\) induces a predictive observation model \(m_\theta(\cdot\mid u)\in\Delta(\mathcal O)\).
Define the predictive KL loss of conjecture \(\theta\) under \(u\) by
\[
    \ell_\theta(u)
    =
    D_{\mathrm{KL}}\!\left(\mu_u\,\|\,m_\theta(\cdot\mid u)\right),
\]
and the set of best predictive conjectures by
\[
    \Theta^\star(u)
    =
    \arg\min_{\theta\in\Theta}\ell_\theta(u).
\]

Following the finite-family likelihood-selection result in Appendix~\ref{app:berk_selection},
the long-run posterior should place mass only on conjectures that are best at predicting the data
generated under \(u\). We therefore define the model-response correspondence
\[
    \Gamma(u)
    =
    \left\{
    w\in\Delta(\Theta):
    \operatorname{supp}(w)\subseteq\Theta^\star(u)
    \right\},
\]
where \(\operatorname{supp}(w)\) is the set of conjectures assigned positive probability. If there is
a unique best predictive conjecture, \(\Gamma(u)\) contains only the point mass on that conjecture;
if several conjectures tie, \(\Gamma(u)\) allows the posterior mass to be split among them.

\paragraph{Definition C.1 (idealized Berk--CVaR equilibrium).}
A pair \((w^\star,u^\star)\in\Delta(\Theta)\times\mathcal U_{\rm id}\) is an idealized Berk--CVaR
equilibrium if
\[
    u^\star\in A(w^\star),
    \qquad
    w^\star\in\Gamma(u^\star).
\]
Equivalently, with \(\mu^\star=\mu_{u^\star}\), the triple
\((w^\star,u^\star,\mu^\star)\) satisfies the closed-loop consistency relation
\[
    w^\star \longrightarrow u^\star \longrightarrow \mu^\star \longrightarrow w^\star .
\]
The conjecture determines the CVaR planning decision, the decision determines the data
distribution, and the data distribution selects the predictive conjectures.

\paragraph{Proposition C.4 (existence under idealized convexity assumptions).}
Suppose:
\begin{enumerate}[leftmargin=1.4em,itemsep=0.1em]
    \item \(\mathcal U_{\rm id}\) is nonempty, compact, and convex, and \(\Delta(\Theta)\) is the simplex
    over a finite model family.
    \item \(\mathcal O\) is finite, \(m_\theta(o\mid u)>0\) for all \((\theta,o,u)\), and
    \(m_\theta(o\mid u)\) is continuous in \(u\).
    \item The true induced observation distribution \(\mu_u(o)\) is continuous in \(u\).
    \item \(J(w,u)\) is continuous in \((w,u)\) and concave in \(u\) for every \(w\).
\end{enumerate}
Then an idealized Berk--CVaR equilibrium exists.

\paragraph{Proof.}
First consider the planning response \(A(w)\). Since \(\mathcal U_{\mathrm{id}}\)  is compact and
\(J(w,u)\) is continuous in \(u\), the maximum theorem implies that
\[
    A(w)=\arg\max_{u\in\mathcal U_{\rm id}}J(w,u)
\]
is nonempty and compact-valued. Since \(J(w,u)\) is concave in \(u\), the argmax set \(A(w)\) is
also convex-valued. Continuity of \(J\) implies that \(A\) is upper hemicontinuous.

Next consider the model response \(\Gamma(u)\). For each \(\theta\in\Theta\),
\[
    \ell_\theta(u)
    =
    D_{\mathrm{KL}}\!\left(\mu_u\,\|\,m_\theta(\cdot\mid u)\right)
    =
    \sum_{o\in\mathcal O}
    \mu_u(o)
    \log
    \frac{\mu_u(o)}{m_\theta(o\mid u)} .
\]
Because \(m_\theta(o\mid u)>0\) and is continuous in \(u\), and because \(\mu_u(o)\) is continuous
in \(u\), each \(\ell_\theta(u)\) is continuous. Since \(\Theta\) is finite,
\(\Theta^\star(u)=\arg\min_{\theta\in\Theta}\ell_\theta(u)\) is nonempty for every \(u\). Therefore
\(\Gamma(u)\) is nonempty, compact, and convex-valued, because it is the simplex over the finite set
\(\Theta^\star(u)\).

We now show that \(\Gamma\) is upper hemicontinuous. Suppose \(u_n\to u\),
\(w_n\in\Gamma(u_n)\), and \(w_n\to w\). If \(w(\theta)>0\), then along a subsequence
\(w_n(\theta)>0\), so \(\theta\in\Theta^\star(u_n)\) along that subsequence. Hence
\[
    \ell_\theta(u_n)
    =
    \min_{\theta'\in\Theta}\ell_{\theta'}(u_n).
\]
Taking limits and using continuity gives
\[
    \ell_\theta(u)
    =
    \min_{\theta'\in\Theta}\ell_{\theta'}(u),
\]
so \(\theta\in\Theta^\star(u)\). Therefore
\(\operatorname{supp}(w)\subseteq\Theta^\star(u)\), and hence \(w\in\Gamma(u)\). Thus \(\Gamma\)
has a closed graph; with compact values and compact range, it is upper hemicontinuous.

Define the correspondence
\[
    T(w,u)=\Gamma(u)\times A(w)
\]
on the compact convex set \(\Delta(\Theta)\times\mathcal U_{\rm id}\). The correspondence \(T\) has
nonempty, compact, convex values and is upper hemicontinuous. By Kakutani's fixed-point theorem,
there exists \((w^\star,u^\star)\) such that
\[
    (w^\star,u^\star)\in T(w^\star,u^\star).
\]
Thus \(w^\star\in\Gamma(u^\star)\) and \(u^\star\in A(w^\star)\), which is precisely an idealized
Berk--CVaR equilibrium.
\hfill\(\square\)

\paragraph{Relation to the implemented planner.}
The implemented \method{} controller is receding-horizon, uses finite scenario samples, and operates
under partial observability and simulator-interface constraints. Proposition~C.4 should therefore be
read only as an idealized consistency statement. It says that, under strong compactness, continuity,
and convexity assumptions, there exists a fixed point linking finite-family predictive model
selection and CVaR planning. It does not imply that the implemented closed-loop robot controller
converges to such a point, nor does it remove model mismatch or real-world hardware effects.

\section{Mechanism Diagnostics for the Representative MuJoCo Seed}
\label{app:mechanism_diagnostics}

\begin{figure}[h]
    \centering
    \includegraphics[width=0.95\textwidth]{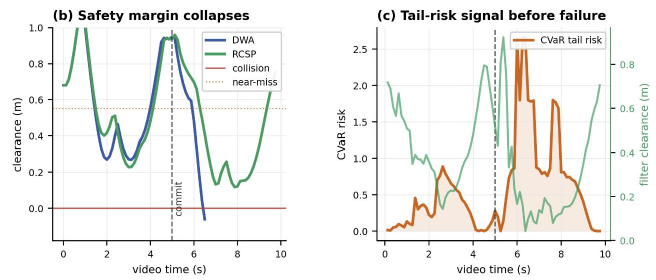}
    \caption{
    Mechanism diagnostics for the representative MuJoCo dynamic-bottleneck seed in
    Figure~\ref{fig:teaser}. Left: the DWA clearance trace collapses after the commitment point and
    crosses the collision boundary, while \method{} maintains positive clearance. Right: the CVaR
    tail-risk signal rises as filtered clearance decreases, suggesting that the planner detects risky
    futures before failure becomes an immediate collision constraint. The diagnostics explain the
    representative rollout and complement the aggregate results in the main text.
    }
    \label{fig:app-mechanism-diagnostics}
\end{figure}

Figure~\ref{fig:app-mechanism-diagnostics} provides a mechanism-level diagnostic for the
representative MuJoCo dynamic-bottleneck seed shown in Figure~\ref{fig:teaser}. The left panel
shows that the DWA clearance trace collapses after the commitment point and crosses the collision
boundary, while \method{} maintains positive clearance and later recovers toward the goal. This
supports the qualitative observation in the rollout frames: the baseline failure is not merely a
last-moment collision event, but the consequence of an earlier commitment into a passage whose
future safety becomes unfavorable.

The right panel visualizes the corresponding tail-risk diagnostic. Around and after the commitment
point, the CVaR risk rises while filtered clearance decreases, indicating that the planner detects
high-risk future interactions before the failure becomes an immediate local collision constraint.
These diagnostics are intended to explain one representative seed; aggregate success, collision,
and transfer results are reported in the main text and the extended tables below.

\section{Extended Experimental Tables}
\label{app:extended_tables}

The following tables provide the full metrics behind the headline results in the main text. They are
not intended as additional independent claims; rather, they document the tradeoffs behind the
controlled-to-public evaluation ladder. We group the results into controlled simulation settings,
where \method{} is expected to help because local progress can hide future tail risk, and transfer
settings, where mature navigation stacks remain strong and the purpose is to expose the boundary of
the method.

\subsection{Controlled Simulation Results}
\label{app:controlled_sim_results}

The controlled simulation results document two different kinds of evidence. In ROS2/Gazebo, the
Nav2+\method{} execution filter supervises official Nav2 velocity commands and reduces dynamic
near-miss failures; this is a deployment-wrapper result, not a full posterior planning result. In
MuJoCo, the dynamic-bottleneck and warehouse-squeeze tasks isolate the intended failure mode more
directly: local or reactive baselines often enter the closing passage, while full posterior \method{}
and fixed-predictor \method{} preserve success and collision avoidance.

\begin{table}[h]
\centering
\caption{ROS2/Gazebo official Nav2 30-seed metrics. The Nav2+\method{} execution filter supervises
official Nav2 velocity commands and improves success, collision, and SPL relative to the official
Nav2 controllers, at the cost of higher safety-cost rate, longer duration, and higher jerk. This row
is a deployment-wrapper test, not a full posterior \method{} planning result. ``Min clear.'' is signed clearance; negative values indicate collision-margin violation.}
\label{tab:app-ros2-full}
\footnotesize
\setlength{\tabcolsep}{3.2pt}
\renewcommand{\arraystretch}{1.08}
\begin{tabular}{lrrrrrrrr}
\toprule
Method & Success & Collision & Safety cost & Min clear. & SPL & Duration & Path length & Jerk \\
\midrule
\textbf{Nav2+\method{} execution filter} & \textbf{0.967} & \textbf{0.033} & 0.428 & \textbf{0.033} & \textbf{0.932} & 31.667 & \textbf{0.927} & 0.112 \\
Nav2 MPPI & 0.533 & 0.467 & 0.408 & 0.027 & 0.533 & \textbf{18.013} & 0.818 & \textbf{0.001} \\
Nav2 DWB & 0.500 & 0.500 & 0.403 & 0.016 & 0.500 & 18.820 & 0.754 & 0.003 \\
Nav2 RPP & 0.467 & 0.533 & 0.405 & 0.013 & 0.467 & 21.650 & 0.802 & 0.003 \\
Nav2 Graceful & 0.333 & 0.667 & \textbf{0.401} & 0.008 & 0.333 & 19.073 & 0.761 & 0.003 \\
\bottomrule
\end{tabular}
\end{table}

\begin{table}[h]
\centering
\caption{MuJoCo robot-navigation ablation summary by environment. These settings isolate dynamic
bottlenecks with differential-drive dynamics, local range observations, and moving obstacles. All
rows use the final all-variant rerun with 30 matched seeds per environment. Full posterior
\method{} uses online posterior-mixture scenario sampling; fixed-predictor \method{} disables the
posterior update while retaining CVaR scoring and the fixed local execution filter.}
\label{tab:app-mujoco-full}
\small
\setlength{\tabcolsep}{6pt}
\renewcommand{\arraystretch}{1.08}
\begin{tabular}{llrrrr}
\toprule
Benchmark & Method & Score & Success & Collision & Safety cost \\
\midrule
BARN/DynaBARN-style & \textbf{Full posterior \method{}} & \textbf{0.464} & \textbf{1.000} & \textbf{0.000} & 17.863 \\
 & Fixed-predictor \method{} & 0.451 & \textbf{1.000} & \textbf{0.000} & 18.297 \\
 & Mean-risk + filter & 0.431 & \textbf{1.000} & \textbf{0.000} & 18.980 \\
 & CVaR-only & -1.166 & 0.167 & 0.467 & 27.648 \\
 & MPC-CBF & -2.163 & 0.067 & 0.900 & 44.215 \\
 & DWA-style & -1.461 & 0.000 & 1.000 & \textbf{15.350} \\
 & TEB-style & -3.333 & 0.000 & 0.967 & 78.767 \\
 & ORCA-style & -3.062 & 0.000 & 1.000 & 68.743 \\
\addlinespace[0.25em]
\midrule
Warehouse squeeze & \textbf{Full posterior \method{}} & \textbf{0.492} & \textbf{1.000} & \textbf{0.000} & \textbf{16.925} \\
 & Fixed-predictor \method{} & 0.480 & \textbf{1.000} & \textbf{0.000} & 17.331 \\
 & Mean-risk + filter & 0.378 & 0.967 & 0.033 & 18.523 \\
 & CVaR-only & -0.995 & 0.200 & 0.467 & 23.173 \\
 & MPC-CBF & -2.247 & 0.000 & 0.933 & 43.573 \\
 & DWA-style & -1.459 & 0.000 & 1.000 & 15.300 \\
 & TEB-style & -2.660 & 0.000 & 1.000 & 55.335 \\
 & ORCA-style & -3.596 & 0.000 & 0.933 & 88.547 \\
\bottomrule
\end{tabular}
\end{table}

The Safety Gymnasium comparison is included as a scope check rather than as the main evidence.
Because the task is more local and more fully observed, CBF-only matches the CVaR+CBF
controller, suggesting that the full posterior scenario layer is most valuable when
dynamic uncertainty and partial observability matter.

\begin{table}[tbp]
\centering
\caption{Safety Gymnasium comparison with trained safe-RL baselines. This benchmark is included
as a supplementary scope check. The first row is a CVaR+CBF controller rather than full posterior
\method{}; CBF-only matches it in this more local and fully observed setting.}
\label{tab:app-safetygym-full}
\small
\setlength{\tabcolsep}{11pt}
\renewcommand{\arraystretch}{1.08}
\begin{tabular}{lrrrr}
\toprule
Method & Eval. episodes & Success & Collision & Safety cost \\
\midrule
CVaR+CBF controller & 20 & 0.950 & 0.050 & 0.650 \\
CBF-only & 20 & 0.950 & 0.050 & 0.650 \\
Goal-PD & 20 & \textbf{1.000} & 0.200 & 3.150 \\
DWA & 20 & 0.450 & \textbf{0.000} & \textbf{0.000} \\
CVaR-only & 20 & 0.650 & 0.150 & 3.100 \\
Trained CPO & 60 & 0.900 & 0.433 & 15.150 \\
Trained PPO-Lag & 60 & 0.983 & 0.517 & 23.100 \\
\bottomrule
\end{tabular}
\end{table}

\subsection{Public Transfer and Boundary Cases}
\label{app:transfer_boundary_results}

The transfer results are deliberately reported as boundary cases. They show that full posterior
\method{} gives better point estimates than its ablations in official DynaBARN/Jackal settings, while
mature navigation stacks retain advantages in strict success, completion, and interface robustness.
This distinction is important for the paper's scope: \method{} is a predictive-risk mechanism, not a
full replacement for a tuned navigation stack.

\begin{table}[h]
\centering
\caption{Paired official DynaBARN statistics comparing fixed-predictor \method{} and full
\method{} on the same 30 generated worlds. Positive values indicate an advantage for full
\method{}; collision and final-distance metrics are written as reductions, so positive values are
also better.}
\label{tab:app-dynabarn-paired}
\small
\setlength{\tabcolsep}{9pt}
\renewcommand{\arraystretch}{1.08}
\begin{tabular}{lrrr}
\toprule
Metric & Mean full-method advantage & 95\% bootstrap CI & $p(\mathrm{adv}\le 0)$ \\
\midrule
Success & 0.167 & $[-0.033,\,0.367]$ & 0.075 \\
Collision reduction & 0.167 & $[-0.033,\,0.367]$ & 0.076 \\
Relaxed completion & 0.033 & $[0.000,\,0.100]$ & 0.356 \\
Min clear & 0.096 & $[-0.156,\,0.337]$ & 0.219 \\
Final distance reduction & 0.222 & $[-0.018,\,0.683]$ & 0.209 \\
\bottomrule
\end{tabular}
\footnotesize
\end{table}

\section{Failure Cases and Limitations}
\label{app:failure_cases}

The results in the main paper are intentionally framed as a targeted mechanism claim rather than
a universal navigation claim. \method{} is designed to help when dynamic obstacles create
predictive near-miss commitments: the robot must decide before the dangerous future becomes an
immediate local collision constraint. The following cases clarify where this mechanism helps, where
it does not yet dominate mature navigation stacks, and what must be improved before stronger
real-world deployment claims can be made.

\paragraph{Public DynaBARN/Jackal transfer.}
Official DynaBARN/Jackal provides the strongest public transfer test in the paper
because it is not tailored to our near-miss environments. On this benchmark, maintained
DWA and TEB achieve higher strict success than full posterior RCSP. This result should
not be read as a contradiction of the main claim. Rather, it shows the boundary between
a predictive-risk planning mechanism and a mature navigation stack with strong
path-following, tuning, and robot-interface robustness. Full posterior/expanded-family
RCSP gives better point estimates than fixed-predictor RCSP, CVaR-only, and MPC-CBF on
endpoint and adapter safety diagnostics, but the full-vs-fixed official-success
advantage is directional and posterior diagnostics are diffuse in this interface. We
therefore use DynaBARN as boundary evidence, not as a decisive posterior-update win.

\paragraph{Fully observed local safety tasks.}
Safety Gymnasium results show that CBF-only can match \method{} on local, fully observed tasks.
This is consistent with the intended scope of the method. Scenario planning is most useful when the
risk is predictive: the relevant danger lies in plausible future moving-obstacle configurations that
are not captured by the current local clearance constraint. When the task is already well described
by instantaneous local safety constraints, the additional scenario layer may provide little benefit.

\paragraph{Latency and smoothness.}
Scenario rollouts and execution filtering introduce additional computation and can make commands
less smooth than mature local controllers. This appears in the ROS2/Gazebo metrics, where the
Nav2+\method{} execution filter improves success and collision outcomes but has higher duration and
jerk. Faster scenario proposal, pruning, parallelization, or learned rollout models are needed to make
the planner more suitable for real-time deployment on physical platforms.

\paragraph{Sensing, model mismatch, and hardware effects.}
The finite-sample interpretation in Appendix~\ref{app:model_conditional_theory} is conditional on
the posterior-weighted scenario model and the quality of the observations. It does not remove model
mismatch, sensing noise, actuation delay, calibration error, wheel--ground interaction, or other
hardware effects. The current paper therefore evaluates \method{} as a controlled simulation study
with public transfer checks, not as a physical-robot deployment claim. Real-robot validation remains
an important next step.